\documentclass[preprints,article,accept,moreauthors]{Definitions/mdpi} 

\graphicspath{{image/}}

\usepackage{fancyhdr}
\usepackage{CJK}

\newcommand{\Lowmodel}[0]{Grouped GCN }
\newcommand{\lowmodel}[0]{grouped GCN }
\newcommand{\Lowmodelabbr}[0]{GGCN }

\newcommand{\highmodel}[0]{multi-linear relationship GCN }
\newcommand{\Highmodel}[0]{Multi-linear relationship GCN }
\newcommand{\Highmodelabbr}{MRGCN }

\newcommand{\BJ}[0]{Beijing }
\newcommand{\SH}[0]{Shanghai }

\usepackage{amsmath}
\usepackage{amsfonts}

\usepackage{graphicx}
\usepackage{algorithm}
\usepackage{makecell}

\usepackage{algorithmic}
\usepackage{multirow}
\firstpage{1} 
\pubvolume{1}
\issuenum{1}
\articlenumber{0}
\pubyear{2022}
\copyrightyear{2022}
\datereceived{} 
\dateaccepted{} 
\datepublished{} 
\hreflink{https://doi.org/} 

\Title{Multi-Modal Graph Interaction for Multi-Graph Convolution Network in Urban Spatiotemporal Forecasting}

\TitleCitation{Multi-Modal Graph Interaction}

\Author{Lingyu Zhang$^{1,3} \dagger$, Xu Geng$^{2} \dagger$, Zhiwei Qin$^{3}$, Yunhai Wang$^{1}$*, Hongjun Wang$^{4}$, Xuan Song$^{4}$, Xiao Wang$^{3}$, Ying Zhang$^{3}$, Jian Liang$^{3}$ and Guobin Wu$^{3}$}

\AuthorNames{Firstname Lastname, Firstname Lastname and Firstname Lastname}

\AuthorCitation{Lastname, F.; Lastname, F.; Lastname, F.}

\address{%
	$^{1}$ \quad Shandong University; \\
	$^{2}$ \quad Hong Kong University of Science and Technology;  \\
	$^{3}$ \quad Didi Chuxing, Beijing; \\
	$^{4}$ \quad Southern University of Science and Technology; \\
} 
	
\corres{Correspondence:cloudseawang@gmail.com; }

\firstnote{This work is finished during  working at Didi AI Labs, Didi Chuxing.} 
\abstract{Graph convolution network based approaches have been recently used to model region-wise relationships in region-level prediction problems in urban computing. Each relationship represents a kind of spatial dependency, like region-wise distance or functional similarity. 
	To incorporate multiple relationships into spatial feature extraction, we define the problem as a multi-modal machine learning problem on multi-graph convolution networks. Leveraging the advantage of multi-modal machine learning, we propose to develop modality interaction mechanisms for this problem, in order to reduce generalization error by reinforcing the learning of multi-modal coordinated representations.
	In this work, we propose two interaction techniques for handling features in lower layers and higher layers respectively. In lower layers, we propose \lowmodel to combine the graph connectivity from different modalities for more complete spatial feature extraction. In higher layers, we adapt multi-linear relationship networks to GCN by exploring the dimension transformation and freezing part of the covariance structure. The adapted approach, called multi-linear relationship GCN, learns more generalized features to overcome the train-test divergence induced by time shifting. We evaluated our model on ride-hailing demand forecasting problem using two real-world datasets. The proposed technique outperforms state-of-the art baselines in terms of prediction accuracy, training efficiency, interpretability and model robustness.}

\begin{document}
\fancyhead{}





	\thispagestyle{fancy} 
\lhead{} 
\chead{} 
\rhead{} 
\lfoot{} 
\cfoot{} 
\rfoot{\thepage} 
\renewcommand{\headrulewidth}{0pt} 
\renewcommand{\footrulewidth}{0pt} 

\section{Introduction}
The deployment of urban sensor networks is one of the most important progresses in urban digitization process. Recent advances in sensor technology enables the collection of a large variety of datasets.
Multi-modality is one of the most significant features in knowledge discovery process in urban computing. Data from different sources are often correlated with each other. For region-level prediction problems, like crowd flow prediction \cite{zhang2016dnn,zhang2017deep} or taxi demand prediction \cite{ke2017short,tong2017simpler,Geng2018SpatiotemporalMC}, it has become a common practice to incorporate a large variety of auxiliary datasets, like weather, POI, road network and events.
In this paper, we define each auxiliary dataset as a modality and study multi-modal learning on multi graph convolution networks (MGCN) for spatiotemporal prediction problems in urban computing. This task is challenging due to complex spatial dependencies and temporal shifting generalization gap. 

Designing spatial feature extraction method is challenging due to complex region-wise spatial dependencies.
GCN-based models \cite{li2018dcrnn_traffic,yao2018modeling} are first used for traffic prediction on road networks. Geng et al. \cite{Geng2018SpatiotemporalMC} proposed Multi-GCN (MGCN) for generic spatiotemporal prediction tasks by stacking three GCNs. Each GCN encodes a unique modality (relationship) of auxiliary data (geo-distance, POI similarity and road network) as graph and extract spatial dependencies from such relationship. 
\begin{figure}[t]
	\centering
	\includegraphics[width=0.9\linewidth]{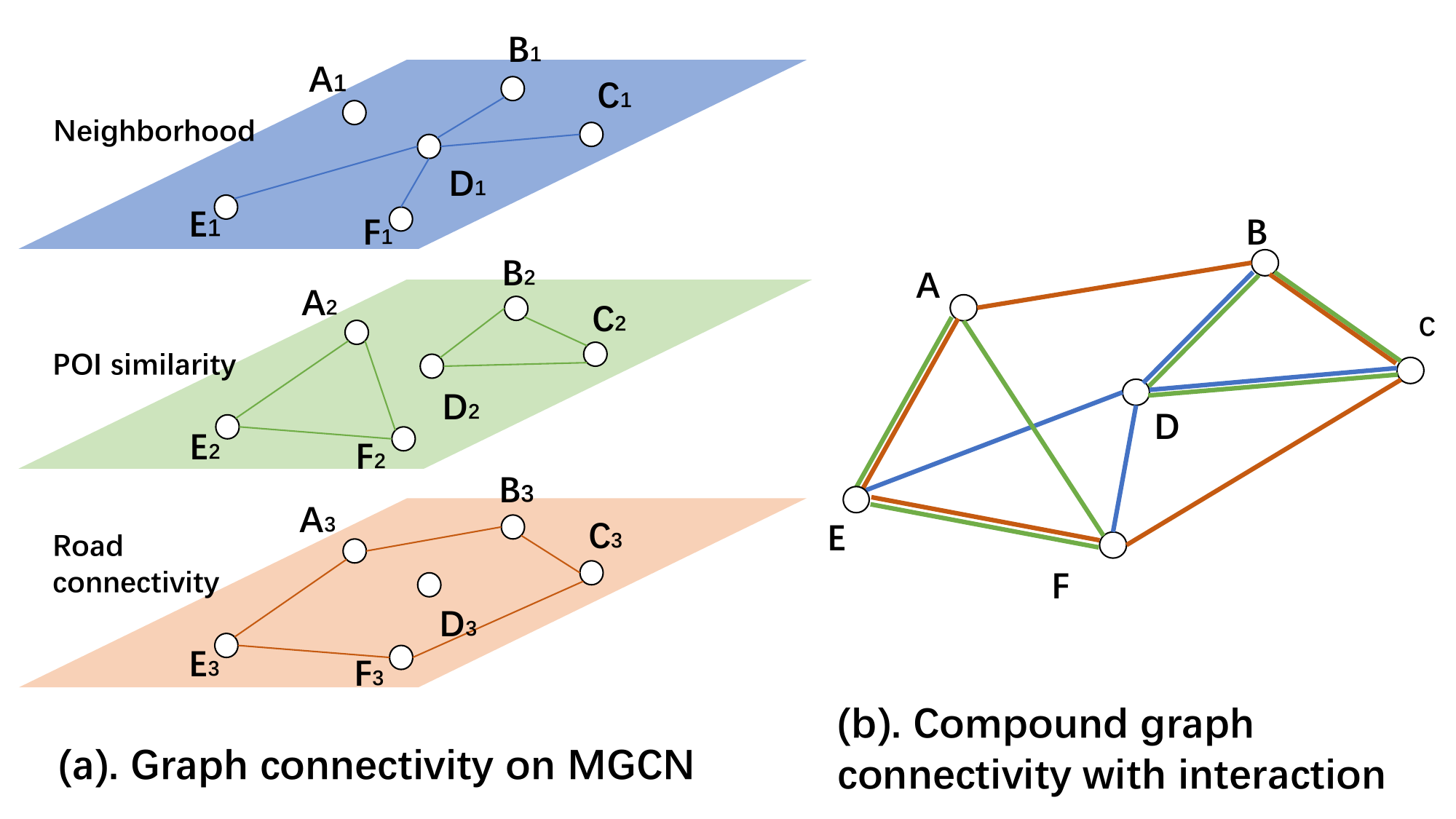}
	\caption{(a) shows graph connectivity for MGCN \cite{Geng2018SpatiotemporalMC} in teach graph. $X_i$ represents vertex (region) X on the i-th graph. Weighted edges between vertices denote region-wise relationship. There is no interaction among graphs. (b) shows compound graph connectivity by adding graph-wise interaction to MGCN. Vertices are connected as long as there exists an edge in any graph.}
	\label{fig:physicalmeaning}
\end{figure}
The spatial feature extraction by MGCN architecture is incomplete, due to the lack of cross-graph connectivities. Figure \ref{fig:physicalmeaning} shows an example for MGCN.
Consider the vertex (region) pair A and D. According to graph topology, A and D are disconnected in all three graphs. MGCN is incapable of extracting features from D for A, or vice versa. However, we argue that the $A-D$ relationship is important. The region pair $A_{3}-B_{3}$ and $B_{2}-D_{2}$ are closely related on road connectivity and POI similarity. A and D are related region pairs for spatial feature extraction. To complete the physical meaning for spatial feature extraction by MGCN, the ideal graph connectivity is shown in \ref{fig:physicalmeaning}(b). It is produced by merging all edges from separate graphs, so that any random walk path is a compound of any kind of relationships. 

Improving model generality to overcome the temporal shifting generalization gap is another challenging task. Temporal pattern for time series data varies along with time. Formally,
\begin{equation*}
    P(X_{t}|X_{t-1},X_{t-2},...)\neq P(X_{t'}|X_{t'-1},X_{t'-2},...) , t\neq t'
\end{equation*}
The gap above defines the divergence between temporal pattern distributions in two different time windows $t$ and $t'$. Such a time shifting gap is often caused by time series fluctuations induced by periodicity, seasonality or miscellaneous factors like weather variation or events. We further discovered that this gap is usually accumulative. A longer temporal interval between two timestamps causes a larger divergence between two distributions. Due to this problem, machine learning models for time series prediction tasks expire frequently. Improving model generality makes the model more robust and avoid of fitting to local time series fluctuations. 

We propose several graph interaction techniques to address to above problem, by enhancing the learning of multi-modal coordinated representations and reinforcing the model performance. Yosinski et al. \cite{yosinski2014transferable} studied feature transferability in deep learning. It shows that features in lower layers are more general and those in higher layers are more specific. According to this phenomenon, we designed two kinds of graph interaction mechanisms correspondingly for lower layers and higher layers.

In lower layers, input spatiotemporal signal maintains its physical properties as engineered features. According to the case in figure \ref{fig:physicalmeaning}, generating latent features via compound graph connectivity makes great sense in terms of spatial feature extraction. 
For lower layer spatial feature extraction, we designed \lowmodel (GGCN), which enables random walk graph convolution on compound graph connectivity. The objective of \Lowmodelabbr is to produce a more abstract multi-modal latent feature representation based on graph convolution operations. 
This technique addresses the first problem on completeness in spatial feature extraction. 

Higher layer features provide high level abstractions for the input signal. It becomes meaningless to explicitly extract feature from a certain region. Leveraging some advances from multi-task learning \cite{zhang2017survey}, we adapt multi-linear relationship learning \cite{long2017learning} to graph convolution networks and try to find shared information among modality-specific representations. 
According to characteristics in GCNs, we propose \highmodel (MRGCN), which imposes tensor normal distribution as the prior distribution of multi-modality graph convolution kernels to learn explainable, robust and fine-grained relationship among modalities. To further enhance the model generality, we propose to freeze part of the covariance structure in the covariance update algorithm, in order to improve output feature independency and alleviate the feature co-adaptation problem. 
The proposed model generates more general high level feature abstractions. This technique also reduce model training time.

On real-world ride-hailing demand data, our model outperforms state-of-the art baselines by a significant margin. Leveraging the advantage of multi-modal and multi-task learning, our model requires less amount of data and time to reach low prediction error. In summary, this paper makes the following contributions:
\begin{itemize}
    \item We propose \lowmodel to produce a compound graph connectivity on multi-modality graph representation. It makes spatial feature extraction on GCN more complete in urban computing.
    \item We propose \highmodel to learn better coordinated representations among modalities. It improves the generality for high level abstractions.
    \item We conduct experiments on two large-scale real-world datasets. The proposed approach achieves more than 10\% error reduction over state-of-the-art baseline methods for ride-hailing demand forecasting.
\end{itemize}

\section{Related Work}
\noindent\textbf{Region-level prediction in urban computing.}
Region-level prediction is a fundamental task in data-driven urban management. There are rich amount of topics, including citizen flow prediction \cite{jiang2021dl,wang2021forecasting,yin2022mtmgnn}, traffic demand prediction \cite{ke2017short,8566163,yao2018deep}, arrival time estimation \cite{li2018kdd_deep_eta} and meteorology forecasting \cite{xingjian2015convolutional,shi2017deep}. For these topics, the region-wise relationships are measured as geographical distance. The spatial structures for these prediction tasks are formulated as regular graphs, which are inherently euclidean structures. Convolution neural networks based models are used for effective prediction.

Non-euclidean structures exist in station-based prediction tasks, including bike-flow prediction \cite{chai2018bike}, traffic volume prediction \cite{li2018dcrnn_traffic,yu2018spatio,yao2018modeling} and point-based taxi demand prediction \cite{tong2017simpler}. The spatial structures for these problems are no longer regular. Graph convolution networks are usually leveraged for spatial feature extraction in these tasks. Non-euclidean structures also exist when incorporating auxiliary data to model region-wise relationships. Yao et al. \cite{yao2018deep} encoded region-wise relationship as a graph and use graph embedding as external features for convolution neural networks. Geng et al. \cite{Geng2018SpatiotemporalMC} used MGCN to model region-wise relationships under multiple modalities. 

\noindent\textbf{Multi-modality in urban computing.}
The core issue for multi-modal machine learning is to build models that can process or relate information from multiple modalities \cite{baltruvsaitis2018multimodal}. Traditional multi-modal machine learning problems focus on human sensory modalities, including audio-visual speech recognition \cite{yuhas1989integration}, multi-media analysis \cite{atrey2010multimodal}
and media description \cite{hodosh2013framing}.
In urban computing, we usually need to harness knowledge from a diverse family of related datasets.
Wei et al.  \cite{wei2016transfer} first categorized the diversity of urban computing datasets, such as POI and air quality as multi-modality and explored feature transferability among different modalities.

Multi-modal fusion is one of the most challenging problems in urban computing.
Most existing works incorporate multi-modality auxiliary data as handcrafted features in a straightforward manner. Tong et al. \cite{tong2017simpler} used multi-modality data as input features for linear regression model. Zhang et al. \cite{zhang2017deep}, Yao et al. \cite{yao2018deep} concatenated auxiliary data to high level abstractions for region-level spatiotemporal prediction networks.

GCN-based approaches encode multi-modality data as region-wise relationships and perform as a static structure in deep learning. The spatial feature extraction process on GCN is associated with these modalities. According to applications in traffic volume prediction \cite{li2018dcrnn_traffic} and taxi demand prediction \cite{Geng2018SpatiotemporalMC}, GCNs are effective in spatial feature extraction on spatial-variant modality data. However, all techniques above fail to build relationship among modalities, which is expected to improve the generality of the learning framework.

\noindent\textbf{Multi-task relationship learning.}
Multi-task relationship learning is a basic approach for multi-task learning. Zhang and Yeung \cite{zhang2012convex} first proposed a regularized multi-task model MTRL by placing a matrix-variate normal prior on model parameter.
\begin{equation*}
    W\sim\mathcal{M}\mathcal{N}(\mathbf{0},\mathbf{\Sigma}_{r},\mathbf{\Sigma}_{c})
\end{equation*}
where $\mathbf{\Sigma}_{r}$ and $\mathbf{\Sigma}_{c}$ are the row and column covariance. Long et al. \cite{long2017learning} proposed Multilinear Relationship Network (MR Network) which learns multilinear relationship on different modes for the joint-task parameter tensor as: 
\begin{gather*}
    W = [W_{1};W_{2};...;W_{t}]\\
    W\sim\mathcal{T}\mathcal{N}_{D_{f}\times D_{c}\times D_{t}}(\mathbf{O},\mathbf{\Sigma}_{f},\mathbf{\Sigma}_{c},\mathbf{\Sigma}_{t})
\end{gather*}
where $W$ refers to the joint weight by concatenating all fully connected weights from all tasks. $D_{f}$,$D_{c}$ and $D_{t}$ denotes to the feature dimension, class dimension and task dimension in the joint weight. $\mathbf{\Sigma}_{f}$,$\mathbf{\Sigma}_{c}$ and $\mathbf{\Sigma}_{t}$ represent covariance for each mode. Experiment results showed that imposing multilinear relationship regularizer on last few fully connected layers in CNN-like structures increased the feature generality and transferability in task specific layers.

However, MR Networks only learn multilinear relationships on fully connected layers. Other deep learning structures, like CNN or GCN have more complicated physical meanings.

\section{Methodology}
\begin{table}
	\centering
\begin{tabular}{l|l|l}
Notation & Type & Meaning \\ \hline\hline
R/V & set   & Set of all regions (vertices)\\
M & set  & Set of all modalities\\
K & scalar & Degree of chebyshev polynomial \\
$I_{d}$ & $\mathbb{R}^{d\times d}$ & Identity matrix with row/column size d  \\
$A_{i}$ & $\mathbb{R}^{|V|\times|V|}$ & Adjacency matrix of $i^{th}$ modality   \\
$L_{i}$ & $\mathbb{R}^{|V|\times|V|}$ & \makecell[l]{Symmetric normalized graph \\ laplacian of $i^{th}$ modality}   \\
$x_{t}$ & $\mathbb{R}^{|V|\times 1}$  & \makecell[l]{A spatiotemporal observation\\ (like ride-hailing demand) value \\at time t}\\
$X^{l}_{j}$ & $\mathbb{R}^{|V|\times f}$ &
\makecell[l]{$f$-dimensional feature of $j^{th}$ \\ modality on $l^{th}$ layer}\\
$O^{l}$ &$\mathbb{R}^{|V|\times 1}$& Output layer as the $l^{th}$ layer\\
$\sigma$ &function& Activation function \\
$f_{1}$, $f_{2}$ & scalar & \makecell[l]{Input feature dimension and \\output feature dimension} \\\hline
\multicolumn{3}{l}{For \lowmodel}\\\hline
$b_{j}$&$\mathbb{R}^{|V|\times f}$& Bias for $j^{th}$ modality\\
$W^l$ & $\mathbb{R}^{\substack{{|M|\times|M|\times} \\ {K\times f_{1}\times f_{2}}}}$ & Weight of $l^{th}$ layer\\
$w^{l}_{i,j}$ & $\mathbb{R}^{K\times f_{1}\times f_{2}}$ & \makecell[l]{Weight for transforming $X^{l}_{i}$\\ to $X^{l+1}_{j}$}\\
$w^{l}_{\alpha}$ &$\mathbb{R}^{f_{1}\times f_{2}}$ & \makecell[l]{Weight corresponding to a specific\\chebyshev polynomial term}\\\hline
\multicolumn{3}{l}{For \highmodel}\\\hline
$|I|,|O|$&scalars& \makecell[l]{Input and Output dimension\\used to measure weight dimension}\\
$W^{l}$&$\mathbb{R}^{|M|\times K \times f_{1}\times f_{2}}$& Weight of $l^{th}$ layer\\
$W^{l}_{i,\alpha}$ &$\mathbb{R}^{f_{1}\times f_{2}}$&\makecell[l]{Weight of $l^{th}$ layer for $i^{th}$ modality\\ and $\alpha^{th}$ chebyshev polynomial term}\\
d & array of 4& dimension of each mode in $\Sigma$\\
$\Sigma_{d_{i}}^{l}$ & $\mathbb{R}^{d_{i}\times d_{i}}$ & Covariance for the $d_{i}^{th}$ mode \\
$\Sigma^{l}$&$\mathbb{R}^{\prod{d_{i}}\times\prod{d_{i}}}$& \makecell[l]{Kronecker decomposable covariance\\ structure for tensor normal distribution}\\
\hline
$L_{high}$&set&\makecell[l]{Higher layers assigned\\ to \highmodel}\\
$L_{low}$&set&\makecell[l]{Lower layers assigned\\ to \lowmodel}\\\hline

\end{tabular}
\caption{Table of notations}
\end{table}

Denote $\mathbb{A} = \{A_{0},A_{1},...,A_{|M|}\}$ as adjacency matrices for different graphs. Each graph corresponds to one of the $|M|$ modalities. In the ride-hailing demand prediction problem, each graph represents a kind of pair-wise spatial relationships for regions, including neighborhood (geo-distance) $A_{N}$, POI similarity $A_{S}$ and road connectivity $A_{C}$ \cite{Geng2018SpatiotemporalMC}.
\begin{align*}
  A_{N,i,j}=&\begin{cases}
               1, \text{if region i and j are adjacent}\\
               0, \text{otherwise}\\
            \end{cases}\\
  A_{S,i,j}=&\text{sim}(P_{v_{i}},P_{v_{j}})\\
  A_{C,i,j}=&\max(0,\text{conn}(v_{i},v_{j})-A_{N,i,j})
\end{align*}
$A_{N}$ defines adjacency relationship between regions. We construct $A_{N}$ by connecting a vertex to its 8 neighbors in a $3\times 3$ grid. 
$A_{S}$ is the cosine similarity between POI vectors of two regions. Each entry in the POI vector represents the number of POIs in a specific category. $A_{C}$ indicates the connectivity between two regions. Two regions are connected as long as there is a highway or subway that directly connects them. 

Define the one-step spatiotemporal prediction task for a certain modality (graph $A_{i}$) on a spatiotemporal observation $x$ as:
\begin{align}
    x_{t} = G(x_{t-1},x_{t-2},...,x_{t-k};A_{i})
\end{align}
where $G$ represents any random-walk based graph convolution network. $x_{t}\in\mathbb{R}^{|V|}$ is the temporal slice of a spatiotemporal observation at time $t$.

When the graph convolution operation $G: (\mathbb{R}^{|V|\times f_{1}};\mathbb{R}^{|V|\times|V|}) \rightarrow\mathbb{R}^{|V|\times f_{2}}$ is defined as polynomial of graph laplacian \footnote{In this work, we use symmetric normalized laplacian:  $L = I-D^{-\frac{1}{2}}AD^{-\frac{1}{2}}$}$L$ with degree up to $K$:
\begin{align}
    G_{W}(X;A) = \sum_{\alpha=0}^{K}L^{\alpha}XW_{\alpha}
\label{eq:ChebNet}
\end{align}
the above definition refers to the graph convolution operation of ChebNet \cite{Bresson2016}. In this work, we use this variation of graph convolution operations.

In multi-modality formulation of this problem, each modality refers to a representation learning process of the same spatiotemporal observation on different graphs.
Following the convention in \cite{baltruvsaitis2018multimodal}, we formalize joint a representation of multi-modality learning problem on multi-graph convolution network as:
\begin{align}
    x_{t} = \mathcal{F}_{A_{i}\in\mathbb{A}}(G_{W}(x_{t-1},x_{t-2},...,x_{t-k};A_{i}))
\end{align}
where $\mathcal{F}_{A\in\mathbb{A}}$ denotes the interaction function across multi-graphs. In previous work \cite{Geng2018SpatiotemporalMC}, it is defined as stacking function in anterior layers and sum function in the output layer.
The major contribution of this work focuses on the design of this interaction function.

\begin{figure}[t]
	\centering
	\includegraphics[width=1\linewidth]{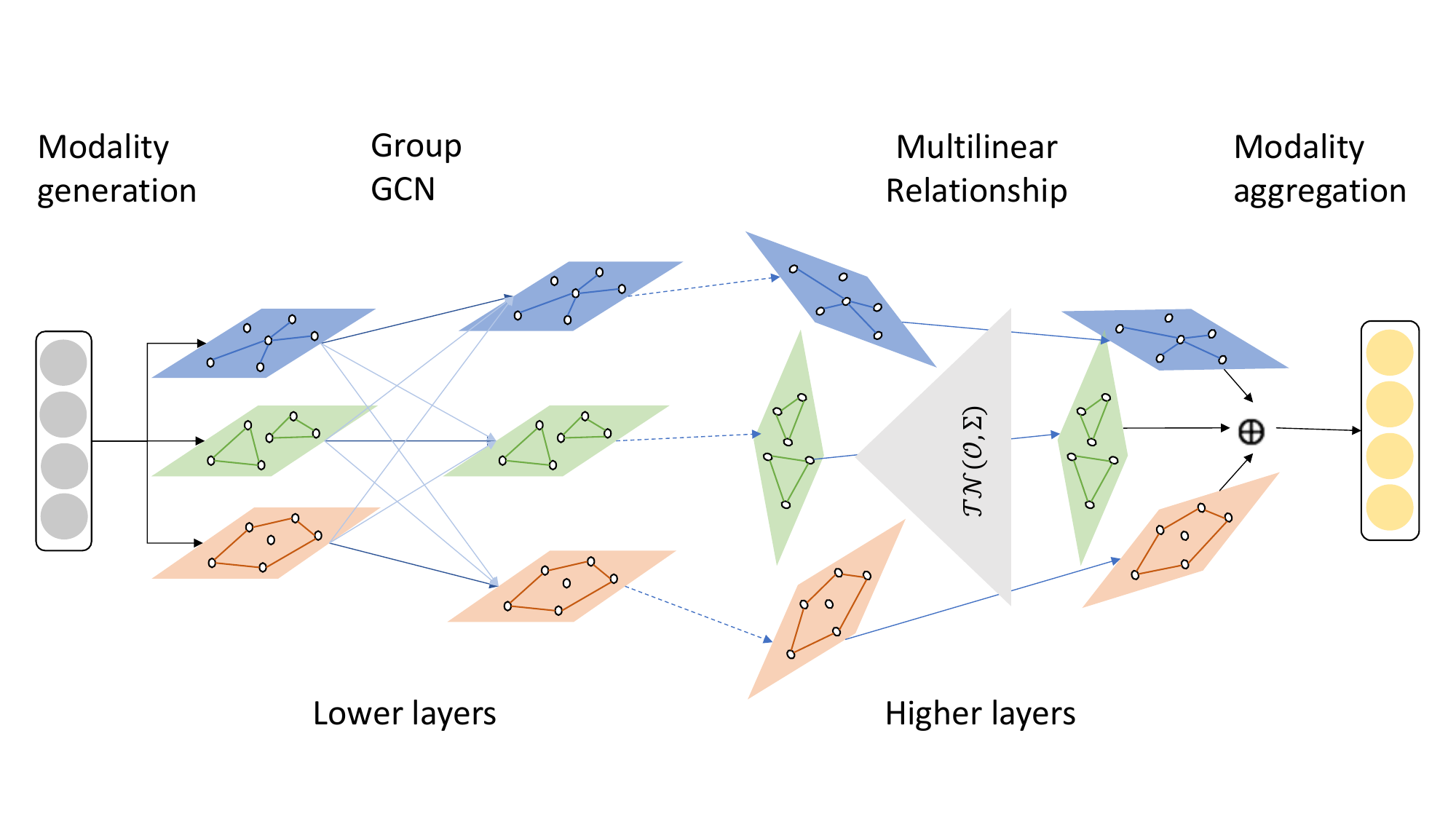}
	\caption{Overview of the proposed graph interaction mechanism for stacked MGCNs. The multi-modality representation of input signals is generated by multi-graphs. In lower layers of deep neural networks, we use \lowmodel to enable inter-graph spatial feature extraction. In higher layers, we use \highmodel to learn modality-wise relationship by imposing tensor normal distribution on the joint representation of parameters. Finally we aggregate modalities to produce output. }
	\label{fig:overview}
\end{figure}

Figure \ref{fig:overview} shows the proposed framework. According to analysis on feature generality \cite{yosinski2014transferable} for deep neural networks, we proposed two techniques for building modality-wise interactions targeted for lower layers and higher layers respectively in stacked MGCNs. In lower layers, the hidden features are concrete. The feature extraction in lower layers are usually general. Considering these facts, we propose to build inter-modality connections to enable inter-graph spatial feature extraction. To distinguish feature extraction parameters, we penalize inter-graph weight and intra-graph weight differently by group regularization. In higher layers, the hidden features are highly abstract that they can no longer maintain their physical properties. Applying inter-modality connections is not applicable. High level features are usually task specific, which is harmful to model generality and transferability. In these layers, we propose to learn multilinear relationship on training parameters of joint modalities, in order to improve the model generality and avoid overfitting the model to local fluctuations.

\subsection*{\Lowmodel}
\begin{figure}[htbp]
	\centering
	\includegraphics[width=1\linewidth]{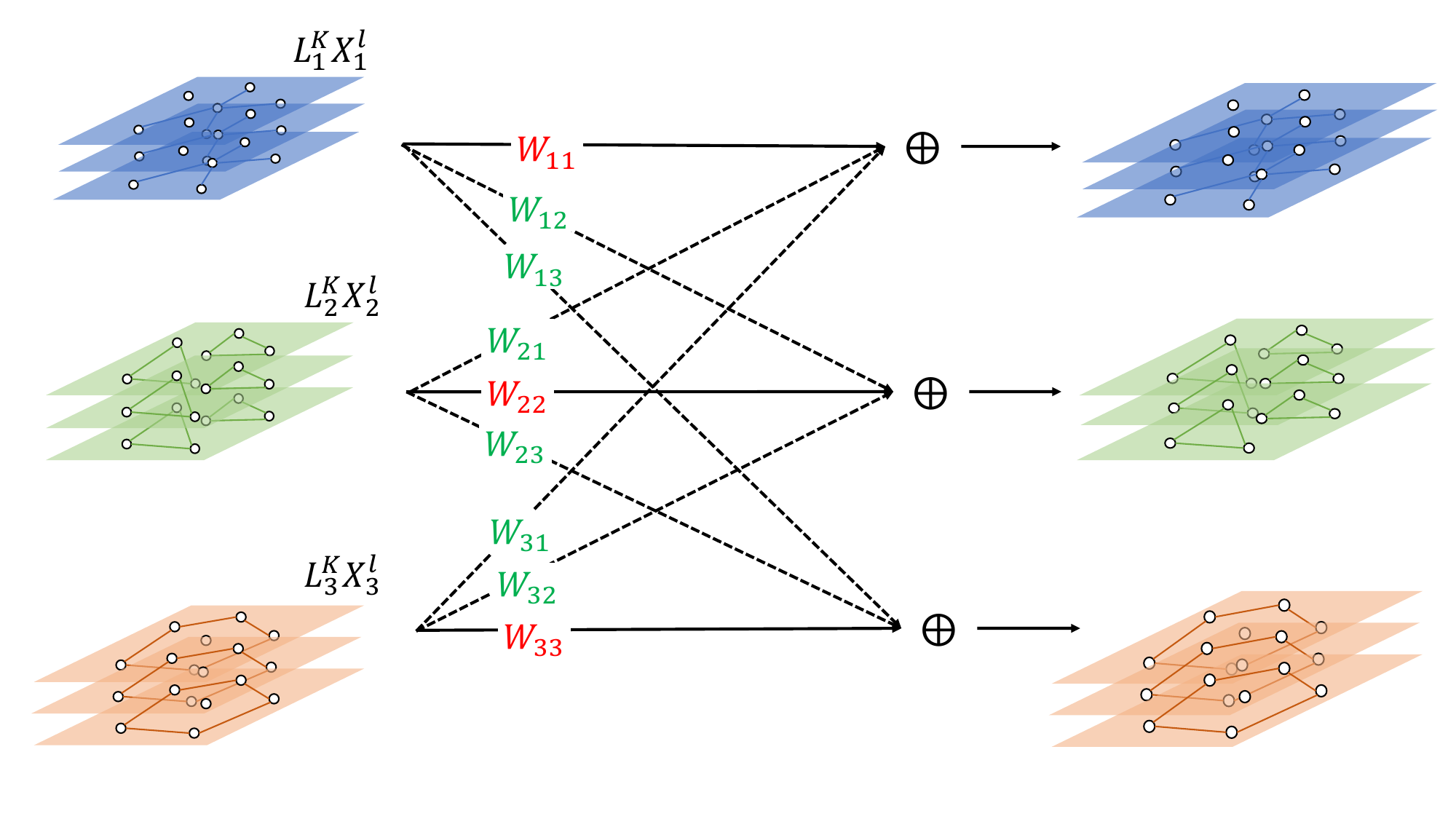}
	\caption{One layer transformation for \lowmodel. Weights marked in red represent intra-modality weights. Green ones represent inter-modality weights.}
	\label{fig:gGCN}
\end{figure}
Figure \ref{fig:gGCN} shows one layer transformation of \lowmodel(\Lowmodelabbr). In lower layers, we use \Lowmodelabbr to build compound graph connectivity, which enables cross-graph spatial feature extraction. 

Denote $L_{i}\in\mathbb{R}^{|V|\times|V|}$ as the graph laplacian matrix of i-th modality. Denote $X^{l}_{i}\in\mathbb{R}^{|V|\times f_{l}}$ as the input signal of the $i^{th}$ modality of the $l^{th}$ layer\footnote{$l,i\in\mathbb{Z^{+}}$}. When $l=1$, $X^{1}_{i}$ represents the raw input and $X^{1}_{i}=X^{1}_{j}, \forall i,j$. 
Define the $l^{th}$ layer parameter $W^{l}$ as:
\begin{align}
    W^{l} = \begin{pmatrix}
w^{l}_{1,1} & w^{l}_{1,2} & ... & w^{l}_{1,|M|}\\
w^{l}_{2,1} & w^{l}_{2,2} & ... & w^{l}_{2,|M|}\\
... & ... & ... & ...\\
w^{l}_{|M|,1} & w^{l}_{|M|,2} & ... & w^{l}_{|M|,|M|}
\end{pmatrix},
\end{align}
Denote $w^{l}_{i,j}\in\mathbb{R}^{f_{l}\times f_{l+1}\times K}$ as the weight matrix to transform the $i^{th}$ modality input to $j^{th}$ modality output via ChebNet transformation $G_{w^{l}_{i,j}}(X_{i}^{l};A_{i})$, where $f_{l}$ and $f_{l+1}$ are the feature dimension of $l^{th}$ and $(l+1)^{th}$ layer. $K$ represents degree of Chebyshev polynomial, which is sliced during the computation of ChebNet. 

The $j^{th}$ modality output is computed as:
\begin{equation}
    X^{l+1}_{j} = \sigma(\sum_{i=1}^{|M|}G_{w_{i,j}^{l}}(X_{i}^{l};A_{i})+b^{l}_{j})
\end{equation}
We denote all weights that transform input to output within same modality, i.e. $w^{l}_{i,j}$ for $\forall i=j$, as intra-modality weights. Similarly, define the inter-modality weight as $w^{l}_{i,j}$ for $\forall i\neq j$.
It's obvious that when all inter-modality weights are set to 0, the graph convolution operation defined above degrades to MGCN. 

Adding cross-modality weights as stated above introduces a tremendous increment on the number of parameters with a factor of $O(|M|)$.
This may boost the model complexity and cause overfitting. To address this issue, we use grouped sparsity \cite{yuan2006model,tibshirani1996regression} to regularize the complexity of parameters. We designed flexible group regularization loss for layer $l$:
\begin{align}
    J^{l}_{1} = \alpha\sum_{i=j}||w^{l}_{i,j}||+\sum_{i\neq j}||w^{l}_{i,j}||
\end{align}
Different from traditional group regularization, we use a tunable parameter $\alpha$ to control the trade-off on penalties for intra-modality weights and inter-modality weights. 
To maintain the difference among modalities, we prefer a smaller $\alpha$ value, in order to introduce less penalty to intra-modality weight. The inter-modality feature extraction focuses on those highly strong relationships. This will help to maintain model generality from multi-modality throughout the proposed \Lowmodelabbr architecture.

The design strategy has several properties that maintain the advantage of GCN models. Firstly, the increment for computational complexity for \Lowmodelabbr is limited. The factor of time complexity increment is $O(M)$, which is polynomial of the number of modalities. In practice, the number of modalities are usually not large. Secondly, the extra computation above to compute intra-modality transformation and inter-modality transformation are naturally independent. It's easy to design a parallel implementation. Finally, \Lowmodelabbr is a linear combination of different graph laplacians, which keep the numerical stability of the original MGCN model when using the normalized symmetric laplacian.

\subsection*{\Highmodel}
In high level layers, latent features no longer maintain their properties as spatiotemporal observations. Instead of building cross-modality connections, we propose to learn multi-linear relationships (MR) on joint-modality weights\footnote{We only keep intra-modality weights in high level layers} by imposing tensor normal distribution as the prior distribution.
\begin{figure}[t]
	\centering
	\includegraphics[width=\linewidth]{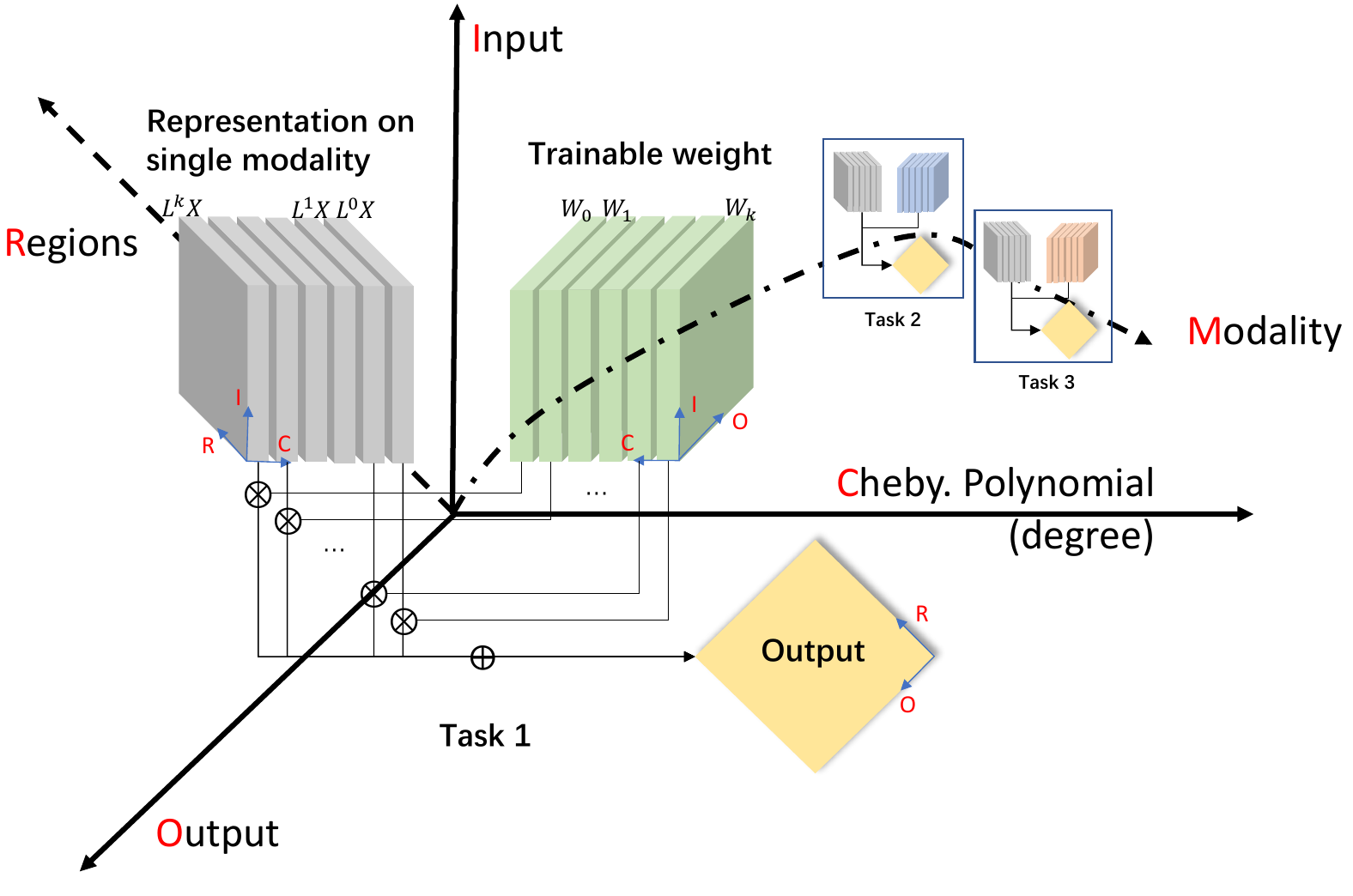}
	\caption{The dimensionality transformation for graph convolution operations in MGCN. Single modality GCN slices input and weight on the $C$ mode, multiply slice pairs and sum up the product.}
	\label{fig:GCN_dim}
\end{figure}

The dimensionality transformation of graph convolution operations in ChebNet is shown in figure \ref{fig:GCN_dim}. There are totally five dimensions in the whole system, including regions/vertices (R, $|R|=|V|$), inputs (I), outputs (O), Chebyshev Polynomial (C, $|C|=K$) and modalities (M).
For each single modality task, the representation of input signals on graph laplacian $L_{i}$ is in three dimensional space of region, input and chebyshev polynomial: $\{L_{i}^{\alpha}X|\alpha=0,1,...,K\}\in\mathbb{R}^{|V|\times|I|\times K}$. The model parameter for the $i^{th}$ modality is in three dimensional space of input, output and chebyshev polynomial: $W^{l}_{i}=\{w^{l}_{i,\alpha}|\alpha=0,1,...,K\}\in\mathbb{R}^{|I|\times|O|\times K}$. The joint representation for multi-modality weight is defined as a four order tensor
\begin{align*}
W^{l} = [W^{l}_{1},W^{l}_{2},...,W^{l}_{|M|}]\in\mathbb{R}^{|I|\times|O|\times K\times|M|}
\end{align*}

Firstly, we impose tensor normal distribution as prior distribution for $W^{l}$
\begin{align}
W^{l}\sim\mathcal{T}\mathcal{N}_{|I|\times|O|\times K\times|M|}(\mathcal{M}^{l},\Sigma^{l})
\end{align}
where $\mathcal{M}^{l}$ is the mean tensor. $\Sigma^{l}=\Sigma^{l}_{I}\otimes\Sigma^{l}_{O}\otimes\Sigma^{l}_{C}\otimes\Sigma^{l}_{M}$ is the kronecker decomposable covariance structure. The density function is estimated as:
\begin{align}
    p(W^{l})=2\pi^{-\frac{\prod_{k=1}^{4}d_{k}}{2}}(\prod^{4}_{k=1}|\Sigma_{k}|^{-\frac{\prod_{k=1}^{4}d_{k}}{2d_{k}}})\times e^{-\frac{1}{2}(W^{l}-\mathcal{M}^{l})^{T}\Sigma^{-1}(W^{l}-\mathcal{M}^{l})}
\end{align}
where $d=[|I|,|O|,K,|M|]$ represents dimensions for each mode, $\Sigma=[\Sigma^{l}_{I},\Sigma^{l}_{O},\Sigma^{l}_{C},\Sigma^{l}_{M}]$. $|\cdot|$ represents determinant. 
According to Long et al.  \cite{long2017learning}, for MAP estimation for model parameters, learning the posterior distribution of $W^{l}$ given training data $(X,Y)$ is equivalent to minimizing the negative logarithm for density of $\prod_{l}P(W^{l})$, where \footnote{Ignores terms irrelevant to $W^{l}$ because they have no gradient during back propagation.}:
\begin{align}
    J^{l}_{2} = \frac{1}{2}(vec(W^{l})^{T}(\Sigma^{l})^{-1}vec(W^{l}))
\end{align}
where $vec(\cdot)$ is the flattening operation to transform a high-dimensional tensor to a 1-d vector.
The flip-flop algorithm for updating covariance matrix of a certain mode $\Sigma_{i}$ is:
\begin{align}
    \Sigma_{d_{i}}^{l} =  \frac{d_{i}}{\prod_{k=1}^{4}d_{k}}(W^{l})_{(i)}(\otimes_{k\neq i}\Sigma_{k})(W^{l})_{(i)}^{T}+\epsilon I_{d_{i}}
\end{align}
where $\epsilon I_{d_{i}}$ is a trade-off term for numerical stability. $(W^{l})_{(i)}$ is the vectorization along the $i^{th}$ mode. Such operation outputs matrix of shape $\mathbb{R}^{(d_{i})\times (\prod_{k\neq i}{d_{k}})}$

We further discovered that the covariance update rule above should not be applied to input (I) and ouput (O) modes. Instead, freezing covariance matrix of input (I) and output (O) mode to identity matrix $I_{d}$ will improve model generality and transferability. 

Observing equation \ref{eq:ChebNet} of ChebNet on the $l^{th}$ layer:
\begin{equation}
        G_{W}(X^{l};A) = \sum_{\alpha=0}^{K}L^{\alpha}X^{l}W_{\alpha}
\end{equation}
where $L^{\alpha}X^{l} \in \mathbb{R}^{|V|\times f_{1}}$, $W\in\mathbb{R}^{f_{1}\times f_{2}}$, usually $|V|>>f_{1}>f_{2}$ due to two facts:
\begin{itemize}
    \item |V| is very large. $L^{\alpha}X^{l}$ is usually sparse.
    \item In higher layers of DNNs, the feature dimension is usually decreasing, i.e. $f_{1}>f_{2}$.
\end{itemize}
According to lemma on matrix multiplication:

\noindent \textbf{Lemma 1.} For matrix multiplication $B=AW$, Rank$(B)$$\leq\min\{ \text{Rank}(A),\text{Rank}(W)\}$
%

The rank for GCN output feature matrix is bounded:
\begin{align*}
    &Rank(G_{W}(X^{l};A))\leq\min\{Rank_{|V|}(L^{\alpha}X^{l}),\\ &Rank_{|f_{1}|}(L^{\alpha}X^{l}),Rank_{|f_{1}|}(W_{\alpha}),Rank_{|f_{2}|}(W_{\alpha})\}
\end{align*}
where $Rank_{|f_{2}|}(W_{\alpha})$ is the rank on $f_{2}$ mode of matrix $W_{\alpha}$. Increasing the $Rank(W_{\alpha})$ on both modes ($f_{1}$ and $f_{2}$) will lift the upper bound of the output rank. It's known that co-adaptation problem \cite{hinton2012improving} limits the generality and transferability of DNNs. Initializing and freezing the covariance matrix along input and output dimension to $I_{I}$ and $I_{O}$, will induce a high rank matrix $W_{\alpha}$, which lifts the upper bound of rank of output features. The inter-neuron dependency is smaller for a high rank output feature matrix, so that the co-adaptation problem is alleviated and model generality is increased.

\subsection*{Multi-modality fusion}
The final layer is the modality fusion layer, in order to aggregate features from different modalities and output a prediction result. For one-step spatiotemporal prediction problem, the output shape is $R^{|V|\times 1}$. The design of modality fusion is straightforward. First, we make sure the last \Highmodelabbr layer reduces the feature dimension to 1. Then, the modality fusion layer is designed as an modality-wise average:
\begin{align*}
    O^{l+1} = \frac{1}{|M|}\sum_{j=1}^{|M|}X^{l}_{j}, X^{l}_{j}\in\mathbb{R}^{|V|\times 1}
\end{align*}

\subsection*{Training algorithm}
We combine all loss functions and summarize it for the entire network:
\begin{align*}
    J_{W}(X,Y) = &\sum_{s\in S}J_{0}(f_{W}(x_{s}),y_{s})+\alpha_{low}\sum_{l\in L_{low}}J_{1}^{l}+\alpha_{high}\sum_{l\in L_{high}}J_{2}^{l}\\
    =& \sum_{s\in S}\frac{1}{|S|}||f_{W}(x_{s})-y_{s}||_{2}\\
    +&\alpha_{low}\sum_{l\in L_{low}}(\alpha\sum_{i=j}^{i,j=1,..,|M|}||W^{l}_{i,j}||+\sum_{i\neq j}^{i,j=1,..,|M|}||W^{l}_{i,j}||)\\
    +&\frac{\alpha_{high}}{2}\sum_{l\in L_{high}}(vec(W^{l})^{T}(\Sigma^{l})^{-1}vec(W^{l}))
\end{align*}
where the $J_{0}$ term is the prediction loss of the model. In this work, we use the rooted mean squared error (RMSE) to measure distance between the predicted value and true value. 
In stacked MGCNs, we set $1,2,...,l_{k}$-th layers to $L_{low}$ and use \Lowmodelabbr to construct graph interactions. The remaining layers $l_{k},l_{k}+1,...$ are set to learn multilinear relationships by MRGCN.
The $J_1$ terms are the \Lowmodelabbr regularizer for each lower layer. The $J_2$ terms are the relationship regularizer for \Highmodelabbr in the higher layers. $\alpha_{low}$ and $\alpha_{high}$ are the trade-off parameters for regularizers.

The overall training algorithm for the entire network, including \Lowmodelabbr and \Highmodelabbr is shown below.

\begin{algorithm}
\caption{Training algorithm for GCN with interactions}
\label{alg:A}
\begin{algorithmic}
\STATE {Set layers $L_{low}=\{1,2,...,l_{k}\}$ to \lowmodel}
\STATE {Set layers $L_{high}=\{l_{k+1},...\}$ to \highmodel}
\STATE {Initialize $\Sigma^{l}_{d}=I_{d}$, $\forall l\in L_{high}$ and $d\in\{|I|,|O|,|C|,|M|\}$}
\STATE {Initialize all weights}
\REPEAT 
\STATE Extract $(x_{i},y_{i})$ from training set as current training batch
\STATE Update model parameter $W$ according to $J_{W}(x_{i},y_{i})$
\STATE Update covariance matrices $\Sigma^{l}_{C}$ and $\Sigma^{l}_{M}$, $\forall l\in L_{high}$
\STATE Evaluate current model using the validation set
\UNTIL{converge (validation error no longer decreases for several epochs)}
\end{algorithmic}
\end{algorithm}

\section{Experiments}
In this section, we compare our graph interaction techniques with state-of-the-art baselines on region-level demand forecasting for ride-hailing service. 

\subsection*{Dataset} 
We conduct our experiments on two real-world large scale ride-hailing datasets collected in two cities: \BJ and \SH. Both of the datasets were collected in main city zone in 2017. We split data to training set (Mar 1st to Jul 31st, 2017), validation set (Aug 1st to Oct 31th, 2017) and test set (Nov 1st to Dec 31st, 2017). The POI data used for $A_{S}$ contains 13 primary categories, including business building, residential building, entertainments, etc. The road network data used for $A_{C}$ is extracted from railway, highway and subway dataset from OpenStreetMap \cite{haklay2008openstreetmap}.

\subsection*{Experiment setting}
The ride-hailing forecasting problem is a one-step spatiotemporal prediction problem to learn predictor $f:\mathbb{R}^{|V|\times T}\rightarrow\mathbb{R}^{|V|\times 1}$. According to previous works \cite{Geng2018SpatiotemporalMC}\cite{ke2017short}\cite{zhang2017deep}, we set T to 5. Physically, it means to predict the ride-hailing demand in the next time interval using the most recent three ones (closeness), the one in the same time yesterday (period) and the one in the same time last week (trend)\cite{zhang2016dnn}. $V$ is the set of regions acquired by partitioning the main city zone to $1km\times 1km$ rectangular grids. Under this setting, there are totally 1296 regions in \BJ and 896 regions in \SH. We set 30 minutes as the time interval for both training data and test data. Each entry in the spatiotemporal tensor represents the number of ride-hailing demand of a certain region in 30 minutes.

We propose a 4-layer MGCN, where the first two layers are \Lowmodelabbr and the last two layers are \Highmodelabbr. The output dimensions for these layers are set to 32,64,32,1. For all graph convolution operations, the max chebyshev polynomial K is set to 4. In \Lowmodelabbr, the tunable $\alpha$ is set to 0.1 to maintain intra-modality properties. In \Highmodelabbr, the trade-off parameter $\epsilon$ is set to $1e-6$. We monitor RMSE on the validation set with early stopping. The regularizers $\alpha_{low}$ and $\alpha_{high}$ are both set to $1e-4$. The neural network is implemented using tensorflow \cite{abadi2016tensorflow} and optimized using adam optimizer \cite{kingma2014adam} with the learning rate as 5e-4 and the batch size as 32. All experiments are conducted on an environment with 10GB RAM and 9GB GPU memory of Tesla P40.

\subsection*{Performance comparison}

\begin{table}[h]
\centering
\setlength{\tabcolsep}{3mm}{\begin{tabular}{c|c|c|c}
\hline
\multicolumn{2}{c|}{Method} & \multirow{2}{*}{\makecell[c]{RMSE in \\ \BJ}} &\multirow{2}{*}{\makecell[c]{RMSE in \\ \SH}} \\
Lower layers           & Higher layers           &   &
\\\hline\hline

\multicolumn{2}{c|}{STMGCN}&10.78& 8.30\\
\multicolumn{2}{c|}{MGCN}& 11.82 & 8.64\\
\multicolumn{2}{c|}{\Lowmodelabbr} & 9.51 & 8.18\\
\multicolumn{2}{c|}{$\Highmodelabbr_{2\Sigma}$} &9.68 & 8.30\\
\Lowmodelabbr& Share weight & 9.59 & 8.13  \\
\Lowmodelabbr& DAN & 9.48 & 8.02 \\
\Lowmodelabbr& $\Highmodelabbr_{4\Sigma}$ & 9.47 & 7.92\\
\Lowmodelabbr& $\Highmodelabbr_{2\Sigma}$ & \textbf{9.31} & \textbf{7.88}\\
\hline
\end{tabular}}
\caption{Experiment performance in \BJ and \SH. The proposed approach achieves best result among all methods}
\label{tab:performance}
\end{table}

\begin{table}[]
\centering

\setlength{\tabcolsep}{2mm}{\begin{tabular}{c|c|c|c}
\hline
\multicolumn{2}{c|}{Method} & \multirow{2}{*}{\makecell[c]{Epoch of \\ converge}} & \multirow{2}{*}{\makecell[c]{Epoch to break \\10.78}}\\
\makecell[c]{Lower layers} &  \makecell[c]{Higher layers}   &                    \\\hline\hline

\multicolumn{2}{c|}{STMGCN}& 115&115\\
\multicolumn{2}{c|}{MGCN} & 110 &    - \\
\multicolumn{2}{c|}{\Lowmodelabbr} & 130 & 55   \\
\multicolumn{2}{c|}{$\Highmodelabbr_{2\Sigma}$} & 95 & 32      \\
\Lowmodelabbr& Share weight & 78 & 32 \\
\Lowmodelabbr& DAN  & 72 &    \textbf{24}   \\
\Lowmodelabbr& $\Highmodelabbr_{4\Sigma}$  & \textbf{51} & 25      \\
\Lowmodelabbr& $\Highmodelabbr_{2\Sigma}$  & 82 & 27      \\
\hline
\end{tabular}}
\caption{Number of epochs required to converge to optima or benchmark. Multi-task-based method reduce training time by at least 50\%. The experiment is done in \BJ dataset.}
\label{tab:speed}
\end{table}
Table \ref{tab:performance} shows experiments comparisons between the proposed methodology, variations and baselines:
\begin{itemize}
    \item MGCN: Use one separate GCN to learn prediction task in each modality. There is no graph interaction among modalities.
    \item STMGCN\cite{Geng2018SpatiotemporalMC}: Use RNN-based model to extract temporal features ahead of MGCN.
    \item Share weight: A common technique in multi-task learning. The GCN weight is shared across modalities in each layer.
    \item Domain adaptation network (DAN)\cite{long2015learning}: Minimizing modality divergence by minimizing cross-modality feature divergence. The divergence used is mean maximum discrepancy (MMD).
    \item $MRGCN_{4\Sigma}$: The proposed \highmodel with all four covariance matrices updated.
    \item $MRGCN_{2\Sigma}$: Proposed method to freeze covariance matrices for input and output coordinates.
\end{itemize}
All proposed methods above are 4-layer MGCNs, with similar hidden feature sizes and same training configurations (learning rate, batch size, etc). We evaluate the model performance according to the prediction error (RMSE) on the test set. The epoch of converge shown in table \ref{table:speed} measures the time consumption for each model to reach its optima. Different models converge to different optima. Achieving a lower error usually costs longer training time. We set the benchmark to 10.78 in \BJ, which is the performance of baseline \cite{Geng2018SpatiotemporalMC} on the same dataset.

The experiments shows following facts. Firstly, according to performance of \Lowmodelabbr, it improves the prediction accuracy for MGCN by invoking more complexity in spatial feature extraction on graphs. With the help of intra-modality transformations, spatial feature extraction is more complete and the model is more expressive. The performance improvement by \Lowmodelabbr is even more significant than incorperating an RNN-based temporal feature extraction process (STMGCN). However, with the increment of the parameter size, the model is more prone to overfitting and requires longer training time.

Secondly, \Highmodelabbr also improves model performance. Compared with GGCN, the influence to prediction error is slightly inferior. There is no significant difference in model capacity and model structure between \Highmodelabbr and MGCN. We infer that multi-linear relationship approach improves prediction performance by improving model generality, so that $\Highmodelabbr_{2\Sigma}$ is less prone to overfit to the local fluctuations in training set and overcomes the gap between training set and test set. Multi-task learning based approaches, including share weight, DAN and \Highmodelabbr all shorten the model training time. Among these approaches, share weight method reduces model complexity by a factor of $O(|M|)$, which brings down the prediction performance. The performance of \Highmodelabbr and DAN are almost the same.

Thirdly, we show that freezing input and output coordinates in \Highmodelabbr is effective. Compared with $\Highmodelabbr_{4\Sigma}$, $\Highmodelabbr_{2\Sigma}$ decreases the prediction error. This validates our assumption that freezing the covariance for input and output dimension on the weight tensor may induce higher independency among neurons, which alleviates the co-adaptation problem, thus improves model generality.

Training speed is another important factor to evaluate machine learning models. Table \ref{table:speed} shows the training time required to achieve the optimal performance of each model. We use the grid search to determine the minimum training length of each model. Given a larger training set than this, the model can't converge to a significantly lower validation error. Compared with the baseline, the proposed method reduces the amount of training set and the length of training time by approximately 50\%.
Among all tested approaches, $MRGCN_{2\Sigma}$ achieves the lowest prediction error on average and on test data after the $4^{th}$ week.
This is an important feature for industrial use. The life cycle for a more generalized model is longer, which reduces the frequency for model update.
\begin{table}[]
	\centering
\begin{tabular}{l|llll}
                    & STMGCN   & GGCN     & MRGCN    & \makecell[c]{GGCN+\\MRGCN} \\\hline
\makecell[c]{Min training \\ length} & 5 months & 5 months & 3 months & 3 months   \\
\makecell[c]{Training time}       & 110 mins & 130 mins & 45 mins  & 60 mins   
\end{tabular}
\caption{Training speed for each model to achieve best performance in ride-hailing demand forecasting task.}
\label{table:speed}
\end{table}

\subsection*{Model generality}
\begin{figure}[htbp]
	\centering
	\includegraphics[width=\linewidth]{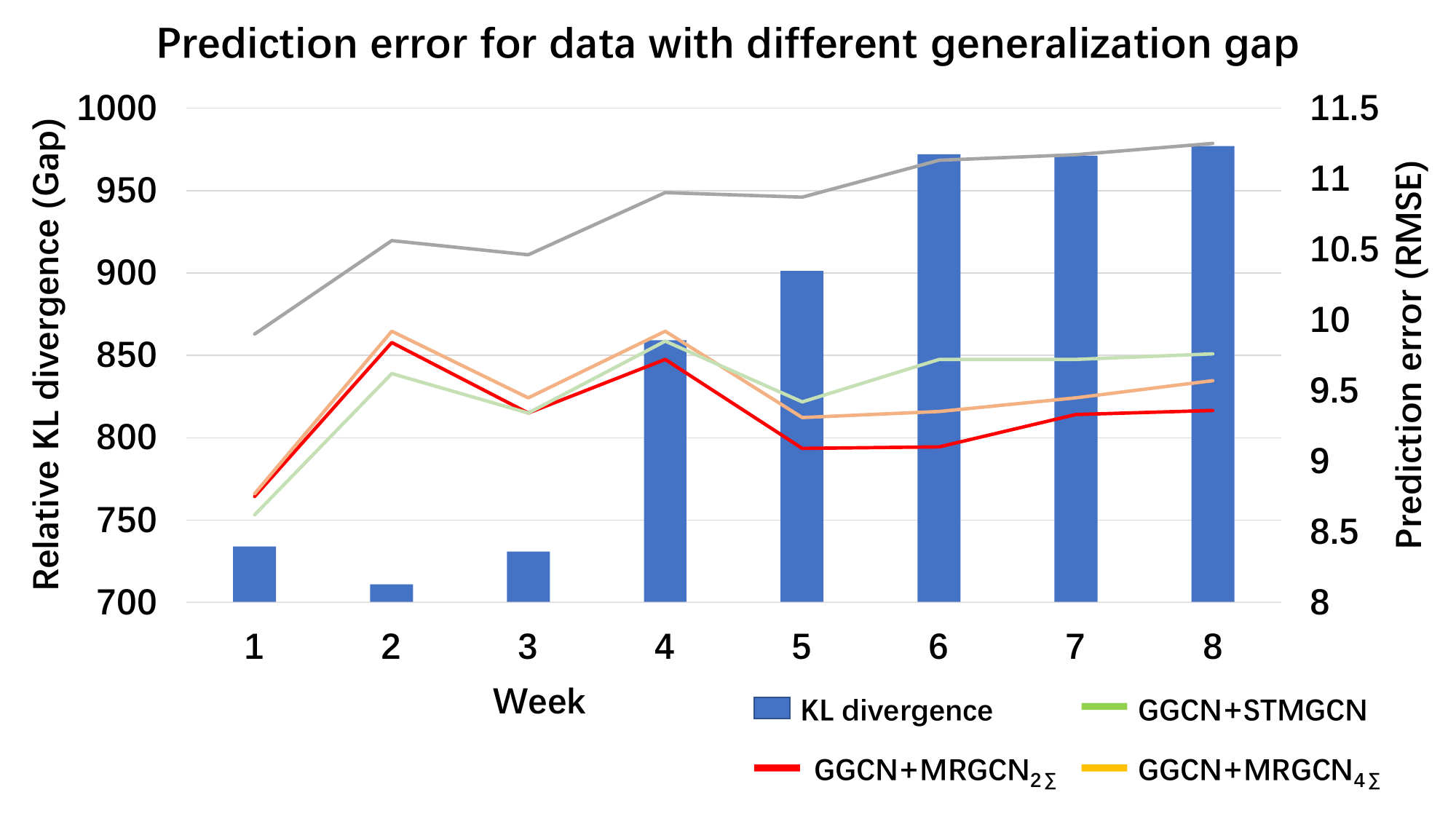}
	\caption{The experiment to test model generality to overcome divergence in temporal data. The relative data divergence in test set accumulates along with time. Multi-task learning based approaches maintains low prediction error when the data divergence is large.}
	\label{fig:generality}
\end{figure}

Figure \ref{fig:generality} shows the generalization ability for different models, which validates above arguments in detail. The data relative divergence (blue bar) is computed as the Kullback Leibler divergence\cite{kullback1951information} between temporal pattern of the last week in training set, and temporal patterns of each week in test set.
We discovered that the gap between training set and test set is accumulative. This indicates that the test data will become more and more divergent from training data with time shifting. Models are expected to be more general to overcome this phenomenon.
According to prediction error by weeks, the prediction error for STMGCN (the gray line) keeps increasing as the test data becomes more divergent. We believe this phenomenon is not caused by model capability, but model generality. For methods including \Lowmodelabbr and MRGCN, the model performance is less influenced by this generalization gap. There is no difference between the model capacity of STMGCN (MGCN) and MRGCN. The network architecture and connectivity are almost the same. This shows that MRGCN has better generalization ability to avoid overfitting to training set.

\begin{figure}[htbp]
	\centering
	\includegraphics[width=0.8\linewidth]{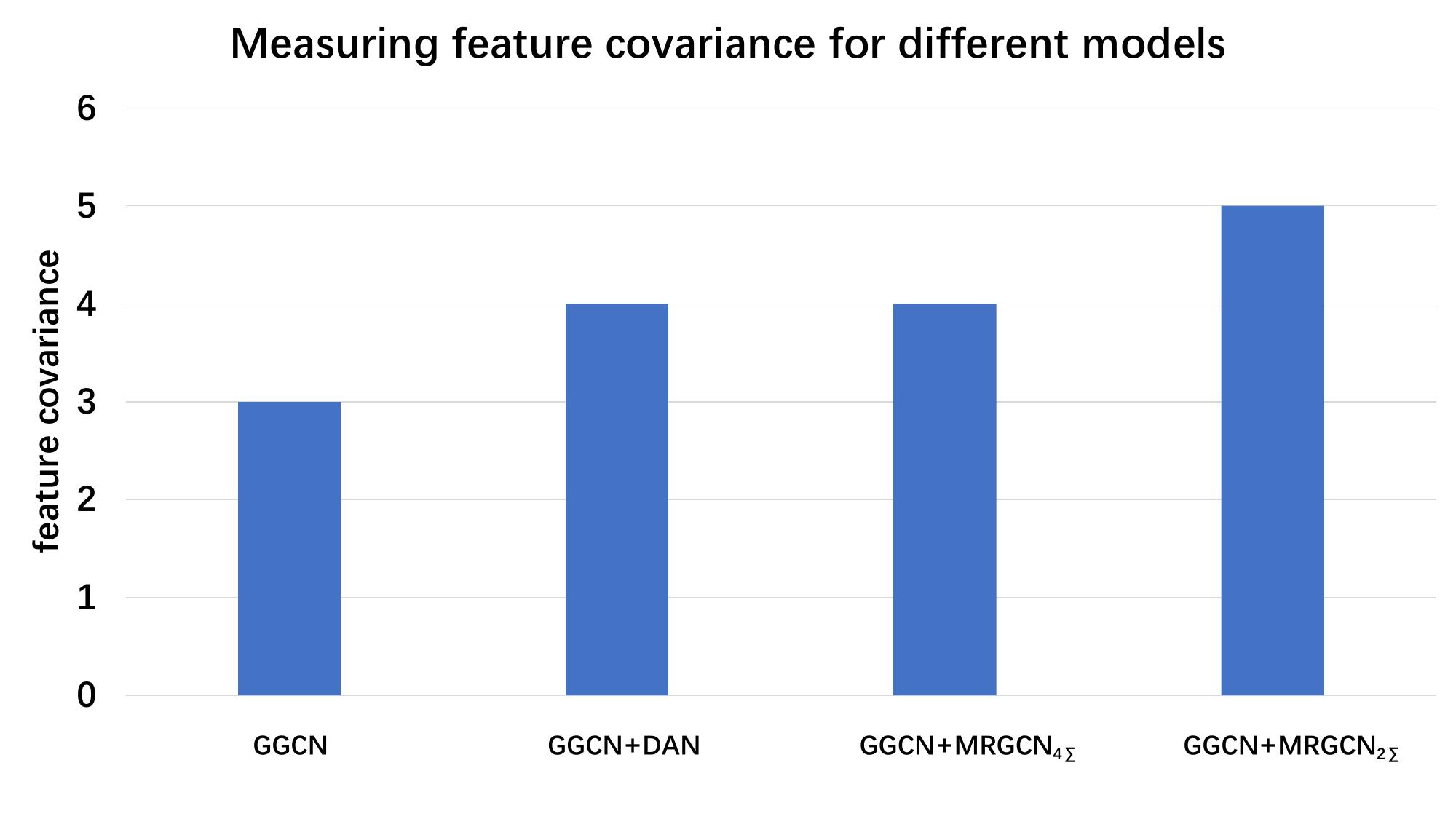}
	\caption{Feature dimension-wise covariance for different models. It's calculated as the negative logarithm of the L2-norm of the covariance matrix of latent features along the feature mode. A higher value indicates higher feature independence.}
	\label{fig:independence}
\end{figure}
Figure \ref{fig:independence} shows the feature inter-dependency of different models. The feature covariance is calculated as negative logarithm of L2-norm of covariance matrix along the feature mode. Feature covariance measures the inter-dependency between different neurons in a hidden layer of deep neural network. A higher value represents a lower absolute value for covariance between neurons and a higher neuron dependency. According to above plot, the neuron independency could be greatly improved by MRGCN. According to Yosinski et al.  \cite{yosinski2014transferable}, co-adapted neurons are the major cause for optimization difficulty in middle layers. Compared with baseline methods, the proposed $\Highmodelabbr_{2\Sigma}$ successfully reduced the coherence among hidden layer units and improved generality and transferability for deep neural networks.

\subsection*{Modality relationship}
\Highmodelabbr learns explainable relationships between modalities by maintaining a modality-wise covariance matrix. In this part, we first show that all modalities are helpful to the learning task. Then, we will explore the relationship between the modality-wise relationship learnt from optimization and relationship between graphs.

\begin{figure}[htbp]
	\centering
	\includegraphics[width=\linewidth]{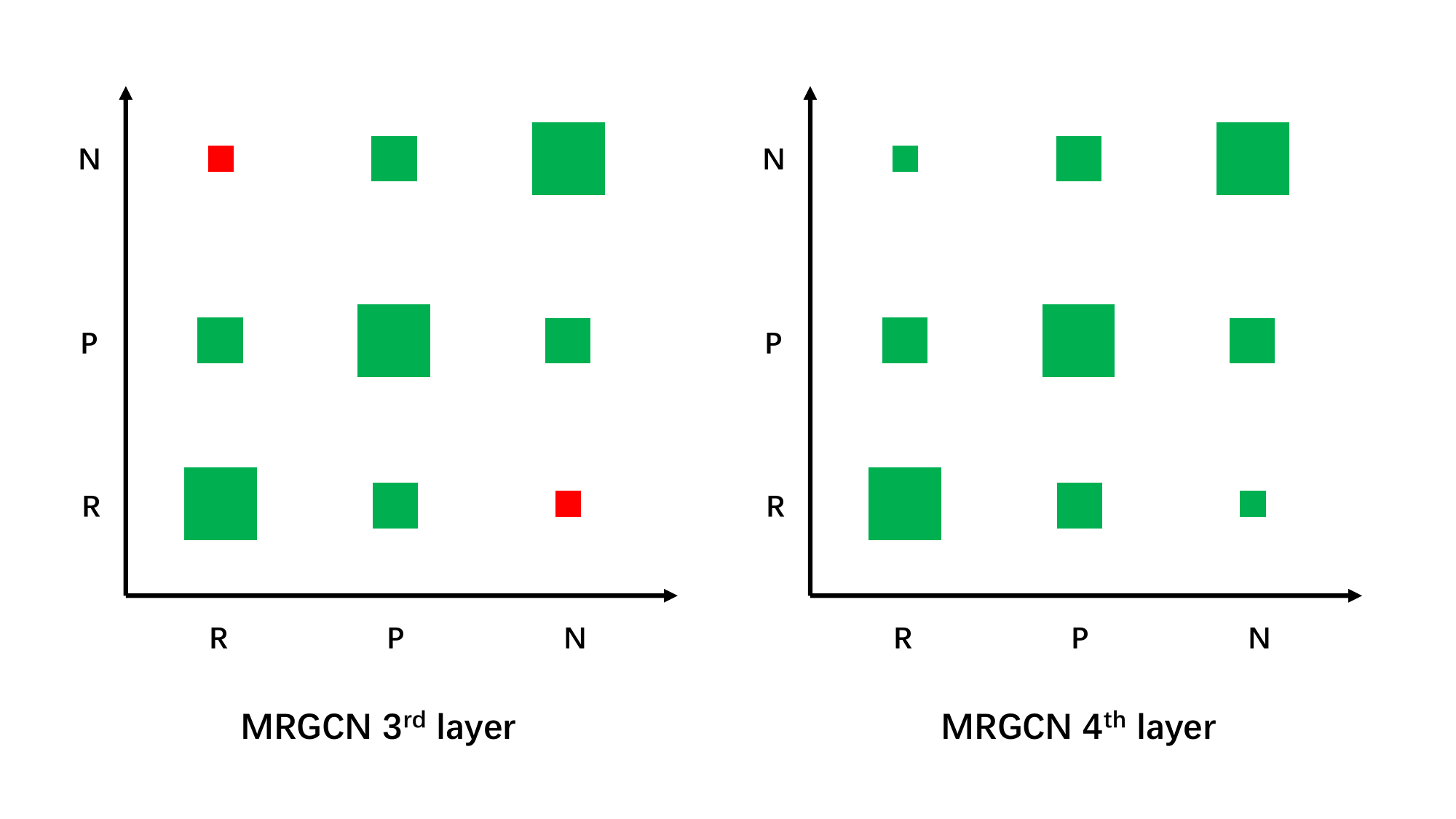}
	\caption{Hinton diagram for modality relationships. The magnitude for relationship is represented by the rectangle size. Green rectangle represents a positive relationship. Red rectangle represents a negative relationship.}
	\label{fig:hinton}
\end{figure}
Figure \ref{fig:hinton} is the Hinton diagram showing the modality-wise relationships for the $3^{rd}$ and $4^{rd}$ layers in \Lowmodelabbr+$\Highmodelabbr_{2\Sigma}$. N, P, R represent modality for Neighborhood $A_{N}$, POI similarity $A_{S}$ and road connectivity $A_{C}$. Similar to the interpretation by \cite{long2017learning}, we could draw several conclusions. Firstly, most of the tasks are positively correlated (green), implying that all modalities could reinforce the learning of others. This conclusion reachs a consensus with ablation study of \cite{Geng2018SpatiotemporalMC} in table \ref{tab:STMGCN}.
\begin{table}[h]
\centering
\begin{tabular}{l|l}
Removed component & RMSE  \\\hline
Neighborhood          & 11.47 \\
POI similarity    & 11.42 \\
Road connectivity & 11.69 \\\hline
ST-MGCN           & 10.78
\end{tabular}
\caption{Ablation study for ST-MGCN. Removing any one modality will result in great damage to the prediction accuracy.}
\label{tab:STMGCN}
\end{table}

Secondly, we discover that the relationship between N and R is weak and random. These two tasks are seemlessly related. Compared with that, the relationship R-P and N-P are stable and robust. We try to explain this phenomenon by comparing the graphs $A_{N}$, $A_{S}$ and  $A_{C}$.
\begin{table}[h]
    \centering
    \begin{tabular}{c|c|c}
         $A_{N}$& $A_{S}$ & $A_{C}$\\\hline
         1.3e-3& 1.4 &1.4e-3
    \end{tabular}
    \caption{The density for each graphs. The graphs are undirected. Density is calculated as $2|E|/|V|(|V|-1)$ }
    \label{tab:density}
\end{table}
\begin{table}[h]
    \centering
    \begin{tabular}{c|c|c|c}
         &$A_{N}-A_{S}$& $A_{S}-A_{C}$ & $A_{C}-A_{N}$\\\hline
        F-measure&0.15&0.17&0\\
        Edit distance &1.1e6&1.1e6&8.8e2
    \end{tabular}
    \caption{Two measurements to show similarity between different graphs. F-measurement considers matched and unmatched edges proportional to graph size. Edit distance measures difference between two edge sets.}
    \label{tab:measure}
\end{table}

Table \ref{tab:density} shows the density for each graph, which measures the connectivity of graph in each modality. According to graph definition, $A_{S}$ is defined as POI similarity between any region pairs, which induces a dense adjacency matrix. $A_{N}$ and $A_{C}$ are sparse. We measure the graph similarity by F-measure and edit distance in table \ref{tab:measure}. According to graph definition, edges in $A_{N}$ are all removed from $A_{C}$, that the edge set $E_{N}\bigcap E_{C}=\emptyset$. From the view of graph connectivity, the prediction task on these modalities are hardly related. The relationship $A_{S}-A_{C}$ and relationship $A_{S}-A_{N}$ are quite similar due to that $A_{S}$ is dense. The analysis above helps to understand figure \ref{fig:hinton}. The relationship between neighborhood (N) and road connectivity (R) is quite random, due to the inherent independency between these two modalities. \Highmodelabbr learns similar modality relationships for similar graph-pairs. The relationship N-P and R-P are maintained to be similar in both layers.
\section{Conclusion and Future Work}
In this work, we propose two graph interaction techniques for multi-modal multi graph convolution networks. We use \Lowmodelabbr in lower layers to complete graph connectivity for better spatial feature extraction by graph convolution networks. In higher layers, we use \Highmodelabbr to learn robust modality relationships. \Highmodelabbr alleviates the co-adaptation problem by lifting the upper bound for feature dependency, thus improves the model generality. The experiment on ride-hailing demand prediction shows that our proposed model outperforms baselines in effectiveness, efficiency and robustness. For the future work, we plan to investigate the following aspects: (1) evaluate the model with other spatial temporal prediction tasks and other region-wise relationships; (2) explore the impact of sparse and dense graphs on this framework; 

\bibliographystyle{ACM-Reference-Format}
\bibliography{bibliography}
\end{document}